\documentclass[conference, a4paper]{IEEEtran}
\IEEEoverridecommandlockouts
\usepackage{graphicx}
\usepackage{float}
\usepackage{stfloats}
\usepackage[caption=false]{subfig}

	\usepackage[turkish]{babel}
	\usepackage[utf8]{inputenc} 
	\usepackage[T1]{fontenc}

\usepackage{cite}
\usepackage[cmex10]{amsmath}
\usepackage{url}

\usepackage{multirow}
\usepackage{array}
\usepackage[table,xcdraw]{xcolor} 

\AtBeginDocument{%
  \renewcommand\tablename{TABLO}
}

\AtBeginDocument{%
  \renewcommand\abstractname{Abstract}
}

\begin{document}


%
\title{Doğal Dil İşlemede Tokenizasyon Standartları ve Ölçümü: Türkçe Üzerinden Büyük Dil Modellerinin Karşılaştırmalı Analizi\\
Tokenization Standards and Evaluation in Natural Language Processing: A Comparative Analysis of Large Language Models on Turkish}

\author{
\IEEEauthorblockN{
M. Ali Bayram\IEEEauthorrefmark{1}, 
Ali Arda Fincan\IEEEauthorrefmark{2}, 
Ahmet Semih Gümüş\IEEEauthorrefmark{2}, 
Sercan Karakaş\IEEEauthorrefmark{3},\\
Banu Diri\IEEEauthorrefmark{1},
Savaş Yıldırım\IEEEauthorrefmark{4}
}
\IEEEauthorblockA{\IEEEauthorrefmark{1}Yıldız Technical University, İstanbul, Turkey\\
Email: malibayram20@gmail.com, diri@yildiz.edu.tr}
\IEEEauthorblockA{\IEEEauthorrefmark{2}Yeditepe University, İstanbul, Turkey\\
Email: ardafincan@icloud.com, ahmetsemih3434@gmail.com}
\IEEEauthorblockA{\IEEEauthorrefmark{3}The University of Chicago, Chicago, IL, USA\\
Email: sercan.karakas@uchicago.edu}
\IEEEauthorblockA{\IEEEauthorrefmark{4}İstanbul Bilgi University, İstanbul, Turkey\\
Email: savasy@gmail.com}
}

\maketitle

\IEEEpubid{\begin{minipage}{\textwidth}
  \vspace{50pt}
  \textbf{979-8-3315-6655-5/25/\$31.00 ©2025 IEEE}
\end{minipage}}

\begin{ozet}
Tokenizasyon, doğal dil işlemede (NLP) büyük dil modellerinin (LLM) dilsel ve anlamsal başarımını doğrudan etkileyen temel bir ön işleme adımıdır. Bu çalışmada, Türkçe gibi morfolojik açıdan zengin ve kaynakları sınırlı dillerin tokenizasyon problemlerini ele alan yeni bir değerlendirme çerçevesi önerilmiştir. Türk eğitim sistemine ait 6.200 çoktan seçmeli sorudan oluşan Türkçe MMLU (TR-MMLU) veri seti kullanılarak, çeşitli tokenizasyon yöntemleri; kelime haznesi, token sayısı, işlem süresi, dile özgü token yüzdesi (\%TR) ve token saflığı (\%Pure) metrikleriyle değerlendirilmiştir. Bu çalışma kapsamında önerilen yeni metrikler, tokenizasyon yöntemlerinin dilsel yapıları koruma yeteneğini ölçmeyi amaçlamaktadır. Yapılan analizlerde, dile özgü token yüzdesinin model başarımı (MMLU skorları) ile token saflığına göre daha yüksek ilişkiye sahip olduğu gösterilmiştir. Ayrıca, büyük dil modellerindeki parametre sayısının yüksekliğinin tek başına daha iyi bir dilsel başarım sağlamadığı; dile özgü tasarlanmış tokenizasyon yöntemlerinin kritik öneme sahip olduğu belirlenmiştir. Önerilen çerçeve, morfolojik açıdan karmaşık diller için güçlü ve uygulanabilir tokenizasyon standartları sunmaktadır.
\end{ozet}

\begin{IEEEanahtar}
Tokenizasyon, Büyük Dil Modelleri (LLM), Doğal Dil İşleme (NLP), Türkçe NLP
\end{IEEEanahtar}

\begin{abstract}
Tokenization is a fundamental preprocessing step in Natural Language Processing (NLP), significantly impacting the capability of large language models (LLMs) to capture linguistic and semantic nuances. This study introduces a novel evaluation framework addressing tokenization challenges specific to morphologically-rich and low-resource languages such as Turkish. Utilizing the Turkish MMLU (TR-MMLU) dataset, comprising 6,200 multiple-choice questions from the Turkish education system, we assessed tokenizers based on vocabulary size, token count, processing time, language-specific token percentages (\%TR), and token purity (\%Pure). These newly proposed metrics measure how effectively tokenizers preserve linguistic structures. Our analysis reveals that language-specific token percentages exhibit a stronger correlation with downstream performance (e.g., MMLU scores) than token purity. Furthermore, increasing model parameters alone does not necessarily enhance linguistic performance, underscoring the importance of tailored, language-specific tokenization methods. The proposed framework establishes robust and practical tokenization standards for morphologically complex languages.
\end{abstract}




%
\IEEEpeerreviewmaketitle

\IEEEpubidadjcol

\section{G{\footnotesize İ}r{\footnotesize İ}ş}

Tokenizasyon, NLP alanında ham metni kelimelere, alt kelimelere veya karakterlere bölerek dil modellerinin girdi formatına uygun hale getiren temel bir ön işleme adımıdır. Bu süreç, modellerin verimliliği ve performansını doğrudan etkilemektedir. Ancak, eklemeli ve morfolojik açıdan zengin dillerde, özellikle Türkçede, tokenizasyonun karmaşıklığı önemli ölçüde artmaktadır \cite{schmidt2024}.

Farklı tokenizasyon yöntemleri üzerine yapılan çalışmalar, bu sürecin NLP modellerinin başarımı üzerindeki etkisini ortaya koymuştur \cite{domingo2019, fujii2023}. Byte Pair Encoding (BPE) \cite{gage1994} ve SentencePiece \cite{kudo2018} gibi yaygın yöntemler, nadir kelimeleri daha iyi temsil etmek için alt kelime seviyesinde segmentasyon sağlamaktadır. Ancak, bu yaklaşımlar, eklemeli dillerin yapısal özelliklerini tam olarak yansıtamamakta ve Türkçeye özgü dil bilgisel kuralları dikkate almamaktadır \cite{koubaa2024}. Türkçede bir kelime, kök ve ekler yoluyla birçok anlam katmanı içerebilir. Yanlış bir tokenizasyon, dilin doğal yapısını bozarak modelin öğrenme sürecine zarar verebilir \cite{samiullah_nodate}.

Bu çalışmada tokenizasyonun başarısı için iki kritik metrik önerilmektedir: \textit{token saflığı} ve \textit{dile özgü token yüzdesi}. Token saflığı, üretilen tokenlerin anlamlı dilsel birimlerle ne kadar uyumlu olduğunu değerlendirirken, dile özgü token yüzdesi, üretilen tokenlerin hedef dilin kelime haznesiyle ne kadar örtüştüğünü ölçmektedir.

\textbf{Motivasyon Örneği:} Türkçedeki \textit{"evlerimizden"} kelimesini ele alındığında; dilbilimsel olarak doğru bir tokenizasyon, kelimeyi \texttt{"ev"}, \texttt{"ler"}, \texttt{"imiz"}, \texttt{"den"} olarak dört bileşene ayırırken, kötü tasarlanmış bir tokenizer bu kelimeyi \texttt{"e"}, \texttt{"vl"}, \texttt{"er"}, \texttt{"imizd"}, \texttt{"en"} gibi anlamsız parçalara bölebilir. Yanlış bölünmüş tokenler, dil modelinin kelimelerin anlam bütünlüğünü korumasını zorlaştırır ve öğrenme sürecini olumsuz etkiler.

Önerilen değerlendirme çerçevesi, Türkçe MMLU (TR-MMLU) veri setini temel alarak farklı tokenizasyon stratejilerini analiz etmekte ve büyük dil modelleri için yeni bir tokenizasyon standardı sunmayı amaçlamaktadır. Bu çalışma, morfolojik olarak karmaşık dillerde tokenizasyonun dilbilimsel tutarlılık açısından nasıl değerlendirileceğini ortaya koyarak, gelecekteki NLP araştırmaları için önemli bir referans oluşturacaktır.
\section{İlg{\footnotesize İ}l{\footnotesize İ} Çal{\footnotesize I}şmalar}

Son yıllarda, farklı diller için çeşitli tokenizasyon stratejileri geliştirilmiş ve bunların dil modellerinin başarımı üzerindeki etkileri incelenmiştir. Bu çalışmalar, dilbilgisel bütünlük, işlem verimliliği ve model ölçeklenebilirliği arasında denge sağlamayı amaçlamaktadır.

\textbf{Dile Özgü Tokenizasyon Çalışmaları:} Düşük kaynaklı ve morfolojik açıdan zengin diller için özelleştirilmiş tokenizasyon yöntemlerinin gerekliliği giderek daha fazla kabul görmektedir. Arapça için geliştirilen \textit{Arabic Tokenizers Leaderboard} \cite{rashad_nodate}, farklı tokenizasyon tekniklerini değerlendirerek Arapça'nın lehçeler arası çeşitliliği ve yazım sistemindeki karmaşıklıkları ele almaktadır. \textit{AraNizer} \cite{koubaa2024} gibi araçlar, BPE ve SentencePiece gibi alt kelime tabanlı yaklaşımları kullanarak Arapça’nın morfolojik yapısını daha iyi yakalamayı ve dil modeli başarımını artırmayı hedeflemektedir. Benzer şekilde, \textit{NbAiLab Tokenizer Benchmark} \cite{rosa2024}, İskandinav dilleri için tokenizasyon stratejilerini değerlendirerek çok dilli modellerin dil başına optimizasyon ihtiyacını vurgulamaktadır. Almanca için yapılan çalışmalar, \texttt{KorAP-Tokenizer} ve \texttt{SoMaJo} gibi araçların yüksek doğruluk oranlarıyla büyük metin veri kümelerinde verimli işlem yapabildiğini göstermektedir \cite{diewald2022}.

\textbf{Türkçe için Tokenizasyon Çalışmaları:} Türkçeye özgü tokenizasyon yöntemleri üzerine yapılan çalışmalar, morfolojik açıdan zengin diller için özelleştirilmiş stratejilere duyulan ihtiyacı ortaya koymaktadır. Erkaya \cite{erkaya2023}, Türkçede alt kelime tokenizasyon yöntemlerinin başarımını detaylı bir şekilde analiz etmiş ve korpus büyüklüğü ile kelime haznesi boyutunun tokenizasyon kalitesine olan etkisini incelemiştir. Bu çalışma, adlandırılmış varlık tanıma, sözcük türü etiketleme, soru yanıtlama ve duygu analizi gibi görevlerde morfolojik olarak optimize edilmiş bir tokenizasyon yaklaşımının performansı artırdığını göstermektedir. Özellikle eklemeli dillerde, morfolojik yapıyı koruyan bir tokenizasyon sürecinin model başarımı açısından kritik olduğu vurgulanmaktadır.

\textbf{Çok Dilli Tokenizasyon ve Ölçeklenebilirlik:} EuroLLM çalışması \cite{martins2024}, büyük kelime hazneleri içeren tokenizasyon modellerinin çok dilli sistemlerde daha başarılı olduğunu göstermektedir. Bu çalışmada, 128.000 alt kelimelik bir BPE tokenizasyon yöntemi kullanılarak düşük kelime başına token oranı (fertility) ve işlem verimliliği sağlanmıştır. Ayrıca, Mistral, LLaMA-3 ve Gemma gibi büyük dil modelleri için yapılan karşılaştırmalar, geniş kelime haznelerinin tokenizasyon doğruluğunu artırdığını ancak hesaplama maliyetlerini yükselttiğini ortaya koymuştur.

\textbf{Verimlilik ve Hesaplama Optimizasyonu:} GitHub tarafından geliştirilen yeni nesil hızlı BPE uygulaması \cite{neubeck2024}, büyük ölçekli veri kümeleri için daha düşük işlem süresi ve daha yüksek verimlilik sağlamaktadır. Rust \textit{et al.} \cite{rust2021}, belirli dillere özel geliştirilen tokenizasyon stratejilerinin çok dilli modellerin başarımını artırdığını göstermektedir. Lin \textit{et al.} \cite{lin_nodate} tarafından önerilen Seçici Dil Modelleme (Selective Language Modeling - SLM) yöntemi, yüksek bilgi içeriğine sahip tokenları önceliklendirerek dil modelinin eğitim sürecini daha verimli hale getirmektedir. Bu yöntem, özellikle Türkçe gibi zengin morfolojik yapılara sahip diller için anlam kayıplarını azaltmada faydalı olabilir.

\section{G{\footnotesize Ö}rev Tan{\footnotesize I}m{\footnotesize I} ve Y{\footnotesize Ö}ntem}

Bu çalışma, eklemeli ve morfolojik olarak zengin dillerde tokenizasyon stratejilerini değerlendirmeyi amaçlamaktadır. Örnek vaka olarak Türkçe seçilmiştir, ancak yöntem, Fince, Macarca ve Uygurca gibi benzer tokenizasyon zorlukları taşıyan diğer dillere uyarlanabilir şekilde tasarlanmıştır.

Değerlendirme, TR-MMLU veri seti \cite{bayram2025} kullanılarak gerçekleştirilmiştir. TR-MMLU, LLM'lerin dilbilimsel ve kavramsal yeteneklerini değerlendirmek için tasarlanmış, 6.200 çoktan seçmeli sorudan oluşan bir benchmark'tır. 67 disiplin ve 800'den fazla konuya yayılan 280.000 soru \cite{bayram_2024_tr_mmlu} arasından seçilerek oluşturulmuştur. TR-MMLU, hukuk, sağlık, tarih ve doğa bilimleri gibi farklı alanları kapsayarak Türkçe'nin morfolojik ve sentaktik kompleksitelerini yansıtmaktadır. 

Bu çalışma, tokenizasyonu hem hesaplama verimliliği hem de dilbilimsel bütünlük açısından değerlendirmek için şu metrikleri kullanmaktadır:

\textbf{Kelime Haznesi Boyutu:} Tokenizer'ın üretebileceği benzersiz token sayısını temsil eder. Daha geniş kelime hazneleri uzun kelime dizilerini daha iyi yakalarken, daha küçük kelime hazneleri aşırı parçalanma riskini taşır.

\textbf{Toplam Token Sayısı:} Bir veri seti işlendiğinde oluşturulan token sayısını ölçer. Düşük token sayısı hesaplama verimliliğini artırabilirken, aşırı parçalanma semantik anlamsızlıklara neden olabilir.

\textbf{Tokenizasyon Süreci:} Tokenizasyonun hesaplama maliyetini belirler. Gerçek zamanlı uygulamalar için hızlı tokenizasyon gereklidir.

\textbf{Dil-Özgül Token Yüzdesi ($\%TR$):} Üretilecek token'ların hedef dilde geçerli sözcük olma oranını gösterir. Tokenizer'ın dille uyumluluğunu değerlendiren kritik bir metriktir.

\textbf{Saf Token Yüzdesi ($\%Pure$):} Token'ların anlamsal ve gramer bütünlüğünü koruyup korumadığını ölçer. Yüksek $\%Pure$ değerleri, kelimelerin temel bileşenlerine sadık kalındığını gösterir.

Bu metriklerin hesaplanmasında, ITU Türkçe NLP Web Servisi \cite{eryigit2014} ve Kalbur kütüphanesi \cite{aksoy2024} kullanılmıştır. Tüm veri setleri, kodlar ve değerlendirme betikleri, Hugging Face ve GitHub platformlarında kamuya açık hale getirilmiştir \cite{bayram2024}. Bu sayede, çalışma özgün, tekrarlanabilir ve farklı diller için genellenebilir bir değerlendirme çerçevesi sunmaktadır.
\section{Deneyler ve Sonu{\footnotesize Ç}lar}

Bu çalışmada, TR-MMLU veri seti kullanılarak dört farklı tokenizer’ın performansı değerlendirilmiştir. TR-MMLU, toplamda 1.605.376 karakter ve 198.193 kelime içeren kapsamlı bir benchmark olup, büyük dil modellerinin geniş bir konu yelpazesinde değerlendirilmesine olanak tanımaktadır \cite{bayram2025}. 

Tablo~\ref{tab:tokenizer-benchmark}, değerlendirilen dört tokenizer'ın karşılaştırmalı sonuçlarını özetlemektedir.

\begin{table}[h!]
\centering
\caption{Tokenizer Benchmark Sonuçları}
\label{tab:tokenizer-benchmark}
\shorthandoff{=}  
\renewcommand{\arraystretch}{1.2} 
\begin{tabular}{|l|c|c|c|c|}
\hline
\rowcolor[HTML]{DDDDDD} 
\textbf{Metrik} & \textbf{gemma-2} & \textbf{llama-3.1} & \textbf{Qwen2.5} & \textbf{aya-expanse} \\ \hline
\rowcolor[HTML]{FFFFDD} 
Model Parametreleri (B) & 27,2 & 70,6 & 7,6 & 32,3 \\ \hline
\rowcolor[HTML]{FFFFDD} 
MMLU Skoru (\%) & 72,10 & 70,42 & 61,68 & 70,66 \\ \hline
Kelime Dağarcığı Boyutu & 256.000 & 128.256 & 151.665 & 255.029 \\ \hline
Token Sayısı & 497.015 & 488.535 & 561.866 & 434.526 \\ \hline
İşlem Süresi (s) & 2,95 & 3,12 & 3,31 & 2,77 \\ \hline
Benzersiz Token Sayısı & 6.383 & 6.823 & 5.752 & 8.562 \\ \hline
TR \% & 48,63 & 45,80 & 40,33 & 50,67 \\ \hline
Pure \% & 37,05 & 30,91 & 30,15 & 32,96 \\ \hline
\end{tabular}
\shorthandon{=}  
\end{table}

Tablo~\ref{tab:tokenizer-benchmark} incelendiğinde, \texttt{gemma-2} modelinin en yüksek MMLU skoru (\%72.10) ve en yüksek Pure \% (\%37.05) değerine ulaştığı görülmektedir. Bu, modelin dilbilgisel olarak tutarlı tokenler üretme yeteneğinin güçlü olduğunu göstermektedir. Aynı zamanda TR \% değeri (\%48.63) ile Türkçeye olan uyumu da dikkat çekicidir.

\texttt{aya-expanse} modeli en yüksek TR \% (\%50.67) değerine ulaşmıştır. Ancak, MMLU benchmark’ında \%70.66 skoru ile güçlü bir performans sergilese de, Pure \% değeri (\%32.96) ile morfolojik olarak bütünlük açısından bazı eksiklikler göstermektedir.

\texttt{llama-3.1}, \%70.42 MMLU skoru ile başarılı bir sonuç elde etmiş ancak düşük Pure \% (\%30.91) değeri, modelin Türkçe morfolojisini tam olarak yakalayamadığını göstermektedir. En küçük model olan \texttt{Qwen2.5} ise en düşük MMLU skoru (\%61.68) ve en düşük TR \% (\%40.33) değerine sahiptir. Bununla birlikte, nispeten küçük kelime dağarcığı (151.665 token) ve düşük işlem süresi (3.31 saniye) sayesinde hesaplama verimliliği sağlamaktadır.

\begin{figure}[h]
    \centering
    \shorthandoff{=}  
    \includegraphics[scale=0.36]{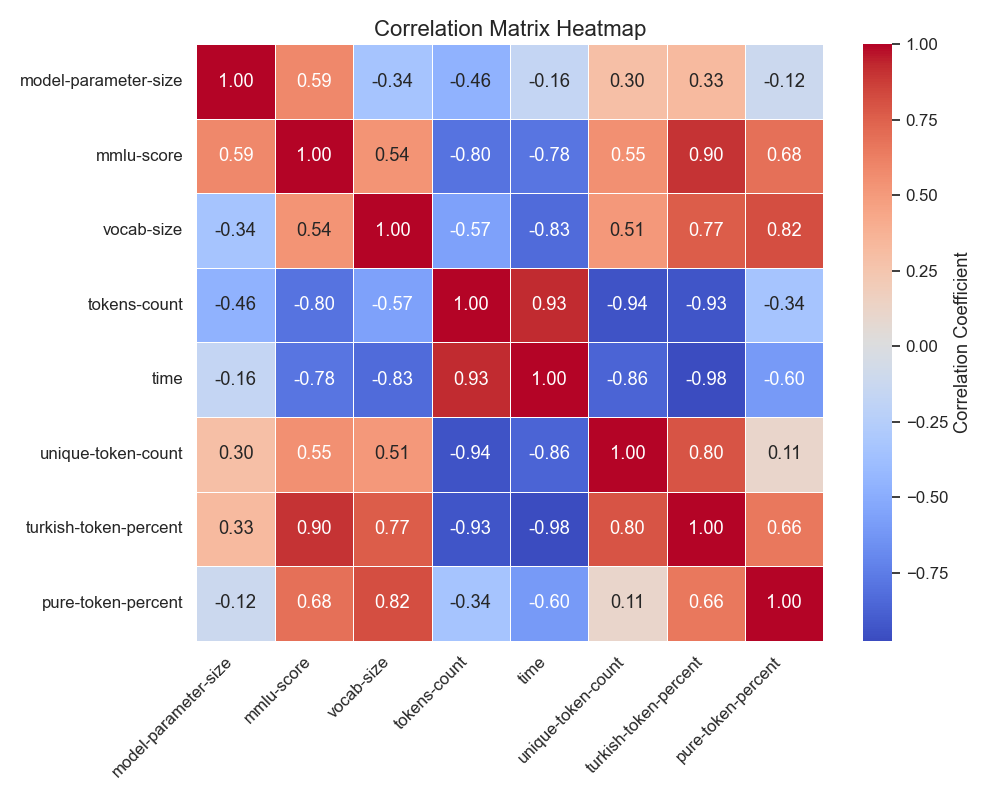}
    \shorthandon{=}  
    \caption{Korelasyon Matrisi: MMLU Skoru, TR \%, Pure \%, Kelime Dağarcığı Boyutu ve İşlem Süresi Arasındaki İlişkiler.}
    \label{fig:correlation_matrix}
\end{figure}

Değerlendirilen metrikler arasındaki ilişkileri daha iyi anlamak için Şekil ~\ref{fig:correlation_matrix}'de ki gibi korelasyon matrisi oluşturulmuştur.

Şekil~\ref{fig:correlation_matrix}, Türkçeye özgü tokenizasyonun model performansı üzerindeki etkisini doğrulayan önemli korelasyonları göstermektedir:

\begin{itemize}
    \item \textbf{TR \% ile MMLU Skoru Arasındaki Korelasyon:} TR \% ile MMLU skoru arasında güçlü bir pozitif korelasyon (\(r = 0,90\)) tespit edilmiştir. Bu, Türkçeye uyumlu tokenizasyonun model performansını artırdığını göstermektedir.
    \item \textbf{Kelime Dağarcığı ve TR \%:} Kelime dağarcığı büyüklüğünün TR \% (\(r = 0,77\)) ve Pure \% (\(r = 0,82\)) ile pozitif korelasyon gösterdiği gözlemlenmiştir. Daha büyük kelime dağarcıkları, tokenizer’ın Türkçe morfolojisine daha iyi uyum sağlamasına yardımcı olmaktadır.
    \item \textbf{Token Sayısı ve İşlem Süresi:} Aşırı token sayısı ve uzun işlem süresi gibi faktörlerin bu metriklerle negatif korelasyon (\(r = -0,93\) ve \(r = -0,60\)) gösterdiği tespit edilmiştir. Bu, aşırı parçalanmış tokenizasyon stratejilerinin dil modelinin bağlam bütünlüğünü bozabileceğini göstermektedir.
\end{itemize}

\begin{figure}[h]
    \centering
    \shorthandoff{=}  
    \includegraphics[scale=0.38]{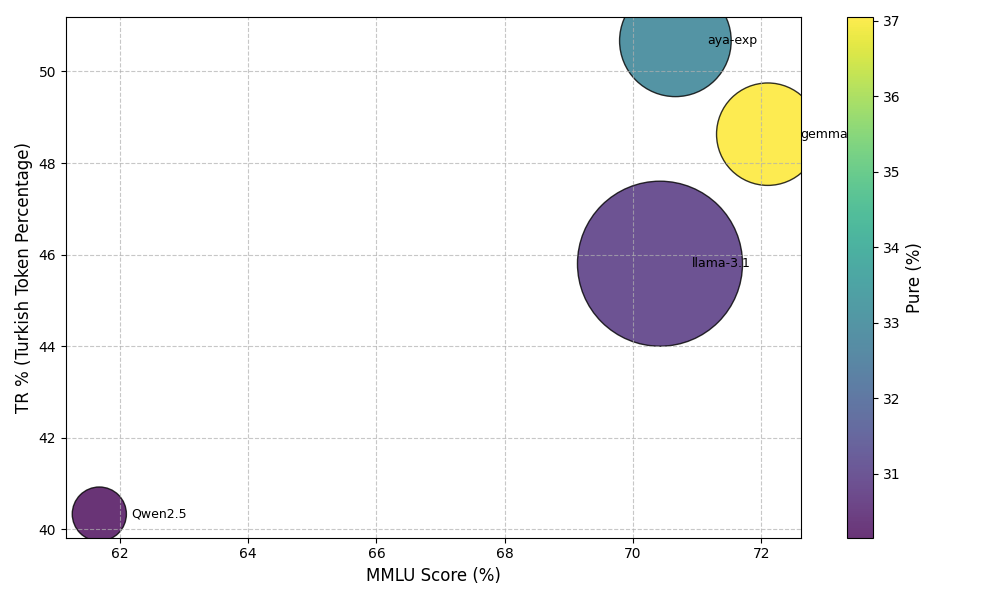}
    \shorthandon{=}  
    \caption{Model Karşılaştırması: MMLU vs TR \%, Parametre Boyutu ve Pure \%.}
    \label{fig:model_comparison}
\end{figure}

Şekil~\ref{fig:model_comparison}, değerlendirilen modellerin MMLU skorlarını TR \% ile karşılaştırmakta ve model parametre boyutlarını temsil eden işaretçi büyüklüğü ile Pure \% değerlerini renk kodlaması ile görselleştirmektedir.

Elde edilen sonuçlar, özellikle morfolojik olarak zengin dillerde, dilbilgisel uyumluluğun model başarısını doğrudan etkilediğini göstermektedir. Dil modellerinin daha yüksek TR \% ve Pure \% değerlerine ulaşması, genel performansı artırmaktadır. Bu sonuçlar, hesaplama verimliliği ile dilbilgisel doğruluk arasında denge kuran tokenizasyon stratejilerinin önemini vurgulamaktadır.
\section{Sonu{\footnotesize Ç}}

Bu çalışma, morfolojik olarak zengin dillerde tokenizasyon stratejilerini değerlendirmek için kapsamlı bir çerçeve sunmuştur. Dilbilimsel bütünlük ile hesaplama verimliliği arasındaki dengenin önemini vurgulayarak, Türkçe gibi eklemeli dillerde etkili tokenizasyonun büyük dil modellerinin başarısını doğrudan etkilediğini göstermiştir. TR-MMLU benchmark'ı kullanılarak gerçekleştirilen analizlerde, token saflığı (\textit{Pure \%}), Türkçe token yüzdesi (\textit{TR \%}), işlem süresi ve kelime dağarcığı büyüklüğü gibi metrikler üzerinden model performansı incelenmiştir.

Elde edilen bulgular, model parametre büyüklüğünün tek başına başarı için belirleyici bir faktör olmadığını göstermektedir. Elde edilen sonuçlar, morfolojik olarak zengin dillerde dilbilgisel uyumu artırmak için özel tokenizasyon stratejilerinin gerekliliğini vurgulamaktadır.

Bu çalışmanın bulguları, yalnızca Türkçe NLP için değil, aynı zamanda düşük kaynaklı ve morfolojik olarak karmaşık diller için de önemli çıkarımlar sunmaktadır. Makine çevirisi, duygu analizi ve bilgi çıkarımı gibi uygulamalarda dilbilimsel bütünlüğü koruyan tokenizasyon stratejileri, model doğruluğunu önemli ölçüde artırabilir. Özellikle tıp ve hukuk gibi alanlarda, özel olarak tasarlanmış tokenizasyon yaklaşımları, ilgili terminolojilere daha iyi uyum sağlayarak, modelin bilgi işleme kapasitesini artırabilir.

Gelecekteki çalışmalar, belirli görevler ve alanlar için dinamik olarak optimize edilebilen tokenizasyon yöntemleri üzerine yoğunlaşmalıdır. Ayrıca, tokenizasyon performansını farklı dillerde karşılaştırarak çapraz-dil çalışmaları yapmak, evrensel ve dile özgü tokenizasyon prensiplerini daha iyi anlamamıza yardımcı olacaktır.

Sonuç olarak, bu çalışma, tokenizasyon stratejilerini değerlendirirken hem dilbilimsel hem de hesaplamalı metrikleri bir araya getiren yeni bir standart sunmaktadır. Özelleştirilmiş tokenizasyon stratejilerinin, küçük ya da daha az optimize edilmiş modellerin bile morfolojik olarak zengin dillerde başarılı olmasını sağlayabileceğini göstermektedir. Bu araştırma, büyük dil modellerinin çok dilli ve alan bazlı NLP görevlerinde daha etkili kullanılmasını sağlamak için geliştirilecek yeni tokenizasyon tekniklerine ışık tutmaktadır.


\begin{thebibliography}{1}

\bibitem{kudo2018}
T. Kudo and J. Richardson, "SentencePiece: A simple and language independent subword tokenizer and detokenizer for Neural Text Processing," \textit{arXiv preprint arXiv:1808.06226}, 2018. [Online]. Available: \url{http://arxiv.org/abs/1808.06226}.

\bibitem{koubaa2024}
A. Koubaa, L. Ghouti, O. Najar, and S. Sebai, "Aranizer: A Custom Tokenizer based on SentencePiece and BPE tailored for Arabic Language Modeling," RIOTU Lab, 2024. [Online]. Available: \url{https://github.com/riotu-lab/aranizer}.

\bibitem{rashad_nodate}
M. Rashad, "Arabic Tokenizers Leaderboard - Hugging Face," Hugging Face Space, [Online]. Available: \url{https://huggingface.co/spaces/MohamedRashad/arabic-tokenizers-leaderboard}. [Accessed: Dec. 13, 2024].

\bibitem{neubeck2024}
A. Neubeck and H. van Antwerpen, "So many tokens, so little time: Introducing a faster, more flexible byte-pair tokenizer," \textit{The GitHub Blog}, Dec. 2024. [Online]. Available: \url{https://github.blog/ai-and-ml/llms/so-many-tokens-so-little-time-introducing-a-faster-more-flexible-byte-pair-tokenizer/}.

\bibitem{diewald2022}
N. Diewald, M. Kupietz, and H. Lüngen, "Tokenizing on scale: Preprocessing large text corpora on the lexical and sentence level," 2022.

\bibitem{rosa2024}
J. de la Rosa and R. Arild, "NbAiLab/tokenizer-benchmark," Nasjonalbiblioteket AI Lab, Nov. 2024. [Online]. Available: \url{https://github.com/NbAiLab/tokenizer-benchmark}.

\bibitem{eryigit2014}
G. Eryiğit, "ITU Turkish NLP Web Service," in \textit{Proceedings of the 14th Conference of the European Chapter of the ACL}, Gothenburg, Sweden, 2014, pp. 1--4.

\bibitem{aksoy2024}
A. Aksoy, "kalbur," Oct. 2024. [Online]. Available: \url{https://github.com/ahmetax/kalbur}.

\bibitem{rust2021}
P. Rust, J. Pfeiffer, I. Vulić, S. Ruder, and I. Gurevych, "How Good is Your Tokenizer? On the Monolingual Performance of Multilingual Language Models," \textit{arXiv preprint arXiv:2012.15613}, 2021. [Online]. Available: \url{http://arxiv.org/abs/2012.15613}.

\bibitem{lin_nodate}
Z. Lin \textit{et al.}, "Not All Tokens Are What You Need for Pretraining," [Online]. Available: \url{https://arxiv.org/abs/2402.06648}.

\bibitem{martins2024}
P. H. Martins \textit{et al.}, "EuroLLM: Multilingual Language Models for Europe," \textit{arXiv preprint arXiv:2409.16235}, Sep. 2024. [Online]. Available: \url{http://arxiv.org/abs/2409.16235}.

\bibitem{gage1994}
P. Gage, "A New Algorithm for Data Compression," 1994. [Online]. Available: \url{http://www.pennelynn.com/Documents/CUJ/HTML/94HTML/19940045.HTM}.

\bibitem{bayram2024}
M. A. Bayram, "tokenizer\_benchmark," Dec. 2024. [Online]. Available: \url{https://github.com/malibayram/tokenizer_benchmark}.

\bibitem{samiullah_nodate}
C. Samiullah, "The Technical User's Introduction to LLM Tokenization," [Online]. Available: \url{https://christophergs.com/blog/understanding-llm-tokenization}. [Accessed: Dec. 22, 2024].

\bibitem{erkaya2023}
E. Erkaya and T. Güngör, "Analysis of Subword Tokenization Approaches for Turkish Language," \textit{2023 31st SIU Conference}, 2023, pp. 1--4.

\bibitem{schmidt2024}
C. W. Schmidt \textit{et al.}, "Tokenization Is More Than Compression," \textit{Proceedings of the 2024 EMNLP Conference}, Miami, Florida, USA, 2024, pp. 678--702.

\bibitem{domingo2019}
M. Domingo \textit{et al.}, "How Much Does Tokenization Affect Neural Machine Translation?," \textit{CICLing 2019}, 2019, pp. 545--554.

\bibitem{fujii2023}
T. Fujii \textit{et al.}, "How do different tokenizers perform on downstream tasks in scriptio continua languages?," \textit{ACL Student Research Workshop}, 2023, pp. 39--49.

\bibitem{bayram2025}
M. A. Bayram \textit{et al.}, "Setting Standards in Turkish NLP: TR-MMLU for Large Language Model Evaluation," \textit{arXiv preprint arXiv:2501.00593}, Jan. 2025. [Online]. Available: \url{http://arxiv.org/abs/2501.00593}.

\bibitem{bayram_2024_tr_mmlu}
M. A. Bayram, "Turkish MMLU: Yapay Zeka ve Akademik Uygulamalar İçin En Kapsamlı ve Özgün Türkçe Veri Seti," \textit{Zenodo}, Aug. 2024, version 1.2. [Online]. Available: \url{https://doi.org/10.5281/zenodo.13378019}.

\end{thebibliography}
\end{document}